\documentclass[letterpaper]{article} 
\usepackage{aaai25}  
\usepackage{times}  
\usepackage{helvet}  
\usepackage{courier}  
\usepackage[hyphens]{url}  
\usepackage{graphicx} 
\usepackage{natbib}  
\usepackage{caption} 
\usepackage{algorithm}
\usepackage{algorithmic}

\usepackage{amsmath}
\usepackage{amssymb}
\usepackage{booktabs}
\usepackage{multirow}
\usepackage{colortbl}
\usepackage{tabularx}
\usepackage[dvipsnames, table]{xcolor}
\usepackage{hyperref}

\usepackage{newfloat}
\usepackage{listings}
\DeclareCaptionStyle{ruled}{labelfont=normalfont,labelsep=colon,strut=off} 
\floatstyle{ruled}
\newfloat{listing}{tb}{lst}{}
\floatname{listing}{Listing}
\title{Differential Alignment for Domain Adaptive Object Detection}
\author{
    Xinyu He\textsuperscript{\rm 1},
    Xinhui Li\textsuperscript{\rm 1},
    Xiaojie Guo\textsuperscript{\rm 1}\thanks{Corresponding author}\\
}
\affiliations{
    \textsuperscript{\rm 1}College of Intelligence and Computing, Tianjin University, Tianjin, China\\

    xy\_he68@tju.edu.cn,
    lixinhui@tju.edu.cn,
    xj.max.guo@gmail.com
}

\begin{document}

\maketitle

\begin{abstract}
Domain adaptive object detection (DAOD) aims to generalize an object detector trained on labeled source-domain data to a target domain without annotations, the core principle of which is \emph{source-target feature alignment}. Typically, existing approaches employ adversarial learning to align the distributions of the source and target domains as a whole, barely considering the varying significance of distinct regions, say instances under different circumstances and foreground \emph{vs} background areas, during feature alignment. To overcome the shortcoming, we investigates a differential feature alignment strategy. Specifically, a prediction-discrepancy feedback instance alignment module (dubbed PDFA) is designed to adaptively assign higher weights to instances of higher teacher-student detection discrepancy, effectively handling heavier domain-specific information. Additionally, an uncertainty-based foreground-oriented image alignment module (UFOA) is proposed to explicitly guide the model to focus more on regions of interest. Extensive experiments on widely-used DAOD datasets together with ablation studies are conducted to demonstrate the efficacy of our proposed method and reveal its superiority over other SOTA alternatives. Our code is available at \url{https://github.com/EstrellaXyu/Differential-Alignment-for-DAOD}.
\end{abstract}

\begin{figure}[!t]
    \centering
    \includegraphics[width=0.96\linewidth]{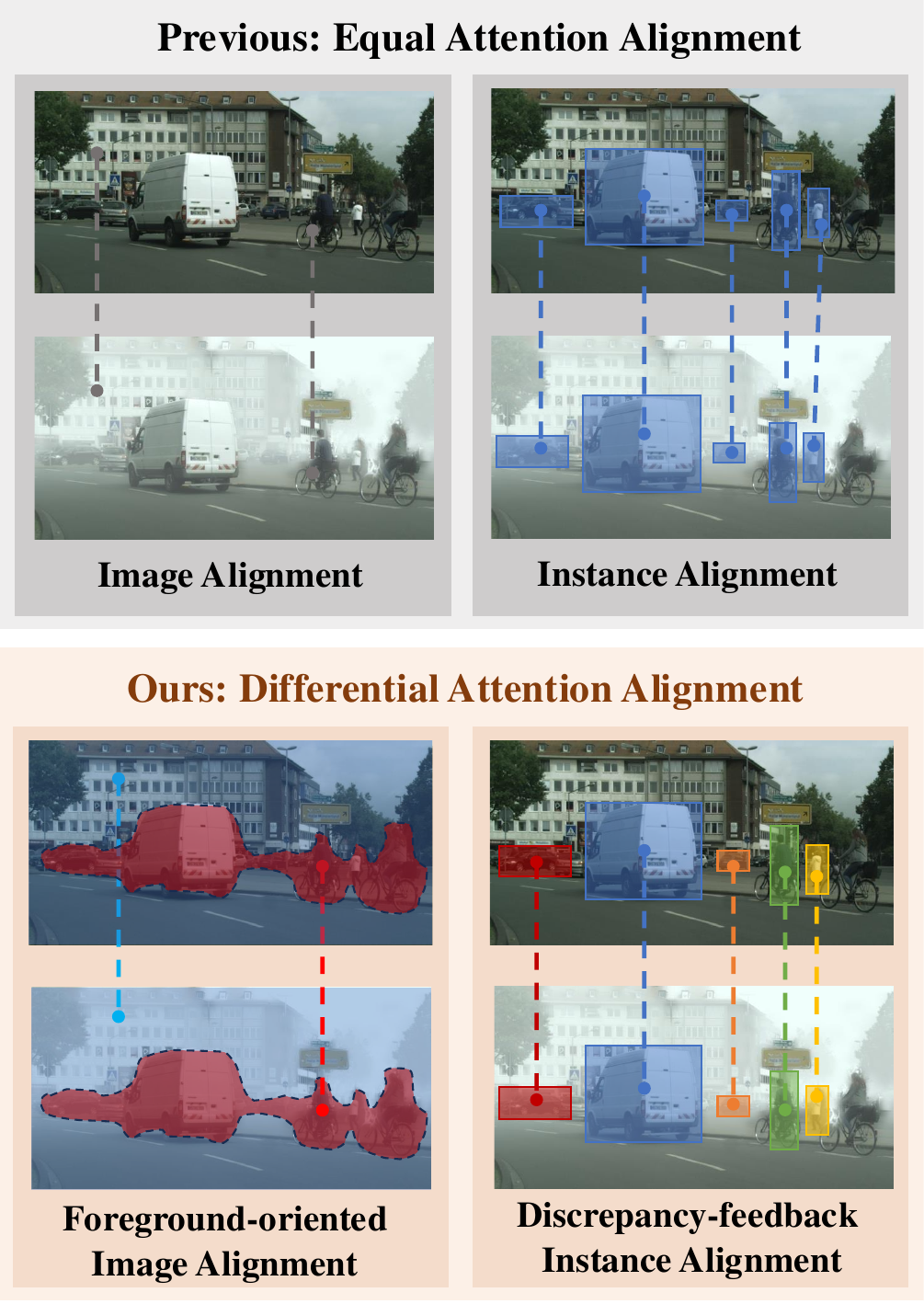}
    \caption{Different from previous methods adopting equal attention feature alignment (\textbf{upper part}), our design manipulates features from the backbone and ROI head with differential attention (\textbf{lower part}). Different colors represent different alignment weights/attentions.}
     \vspace{-5mm}
    \label{fig:compare}
\end{figure}

\section{Introduction}
As one of the fundamental tasks in computer vision, object detection plays a vital role in a wide spectrum of applications, such as autonomous driving~\cite{48odautodriving}, video surveillance~\cite{49surveillance}, and person re-identification~\cite{50reindentification}, to name a few. With the significant development of deep learning and the availability of large-scale annotated datasets, it has experienced remarkable performance improvements in recent years~\cite{45fastrcnn, 1fasterrcnn, 46yolo, 47detr, 67rtdetr}. 
Despite such advancements, the detection environment is not always as expected in practice, leading to performance discrepancies between the training and testing domains, \textit{a.k.a. the domain shift issue}~\cite{4Da-faster, 13sfa}. This shift, driven by variations in illumination, weather, background, and other factors, frequently degrades the performance of detection.
To mitigate this issue, a straightforward way is to acquire and annotate ample and diverse real-world data for training. Nevertheless, even if this manner were possible, it would be extremely laborious and time-consuming.

As a more appealing option, domain adaptive object detection (DAOD)~\cite{11AT, 20mtor, 3PT, 13sfa, 16aqt, 17mrt, 4Da-faster} has emerged to solve the challenge by following the principle of source-target feature alignment. 
The goal is to generalize the model trained on labeled source-domain data to the unlabeled target domain, thus alleviating the reliance on target-domain annotations.
In recent years, several methods~\cite{4Da-faster, 11AT,13sfa, 16aqt, 17mrt} have been proposed in the literature, which employ adversarial learning to align the distributions of the source and target domains as a whole.
Despite having made advancements, they hardly account for \emph{the varying importance of different regions} in feature alignment. For example, foreground objects like cars and persons should be prior to background elements such as roads and sky; and instances in different haze densities should be treated unequally during alignment. Thus, it is desired to design a \emph{differential alignment} mechanism that can effectively adjust attention to regions of different importance, instead of treating all the regions uniformly. 

To achieve the goal, we delve into the differential alignment strategy via investigating cues to prioritize the treatment of critical features during the model adaptation process. This study presents two modules to respectively cope with (1) instances under different circumstances, and (2) foreground \emph{vs} background areas. More concretely, a Prediction-Discrepancy Feedback instance-level Alignment module, namely PDFA, is proposed to dynamically adjust the model's attention on different instances. It can automatically assess the amount of domain-specific information contained in a given instance, which in turn determines the level of alignment effort required for that instance.
In addition, an Uncertainty-based Foreground-Oriented image-level Alignment module (termed UFOA) is customized to explicitly guide the model to concentrate more on foreground regions than background ones. 
In other words, UFOA can sense regions of higher interest/importance and concern more on such regions than the others to overcome the deficiency of traditional feature alignment, \emph{i.e.}, identical attention on foreground and background features. Please see Fig.~\ref{fig:compare} for illustration. Through these two designs, our method can effectively address the varying importance of different regions in feature alignment with significant performance gains over other SOTA alternatives.

Our primary contributions can be summarized as follows:
\begin{itemize}
    \item We propose an instance-level alignment module (PDFA) to dynamically adjust the focus of model based on the prediction-discrepancy feedback, which enables differential alignment of different instances according to their richness in domain-specific information.
    \item We develop an image-level alignment module (UFOA) to adaptively prioritize foreground-object areas, which mitigates the limitation of traditional equal feature alignment by balancing foreground and background through an uncertainty factor.
    \item Extensive experiments are conducted to verify the efficacy of our design, and show that our proposed method remarkably outperforms existing approaches by more than 4$\%$ in AP$_{50}$ on three DAOD benchmarks. 
\end{itemize}

\section{Related Work}
In recent years, various DAOD schemes have been devised to alleviate the performance degradation when applying models trained on a source domain to data from another domain with different distributions. 
Existing methods can be roughly grouped into three classes, \emph{i.e.}, domain translation, self-training, and domain alignment approaches. Domain translation methods realize the adaption purpose by transforming the source-domain data to resemble the target domain, typically adopting generative models. Although valid in aligning distributions to some extent, these schemes are fundamentally constrained by information loss, and substantial computational requirement, limiting their applicability and performance in comparison with approaches in the other two categories. Given the focus of this work, we will primarily review representative methods in self-training and domain alignment families.

\textbf{Self-Training with Pseudo Labels.} 
Methods in this group, typically built upon teacher-student architectures, utilize pseudo labels for unlabeled data, whose effectiveness has been witnessed by several works. 
To be specific, AT~\cite{11AT} arms the self-training paradigm with weak-strong data augmentation, yielding noticeable increase in accuracy.
While PT~\cite{3PT} further leverage the uncertainty of pseudo-labels to promote adaptation during training. 
Alternatively, new advances in general object detection have also promoted the development of DAOD.
As a consequence, MTTrans~\cite{14mttrans} constructs an end-to-end cross-domain detection Transformer based on the teacher-student framework. 
From the perspective of capturing context relations within target-domain images, MIC~\cite{12mic} and MRT~\cite{17mrt} introduce the masked image consistency into the self-training framework to generate high-quality pseudo-labels. Besides, HT~\cite{62ht} enhances the quality of pseudo labels via regularizing the consistency of classification and localization scores.
Although significant progress has been made, the aforementioned methods update the student model via indiscriminately reducing the discrepancy between predictions by the teacher and the student, which overlook meaningful information conveyed by different degrees of discrepancy. That is to say, greater discrepancy of prediction shall reflect heavier domain-specific information existed in corresponding regions. Our approach harnesses this potential by designing a prediction-discrepancy feedback mechanism.

\begin{figure*}[t]
    \centering
    \includegraphics[width=\textwidth]{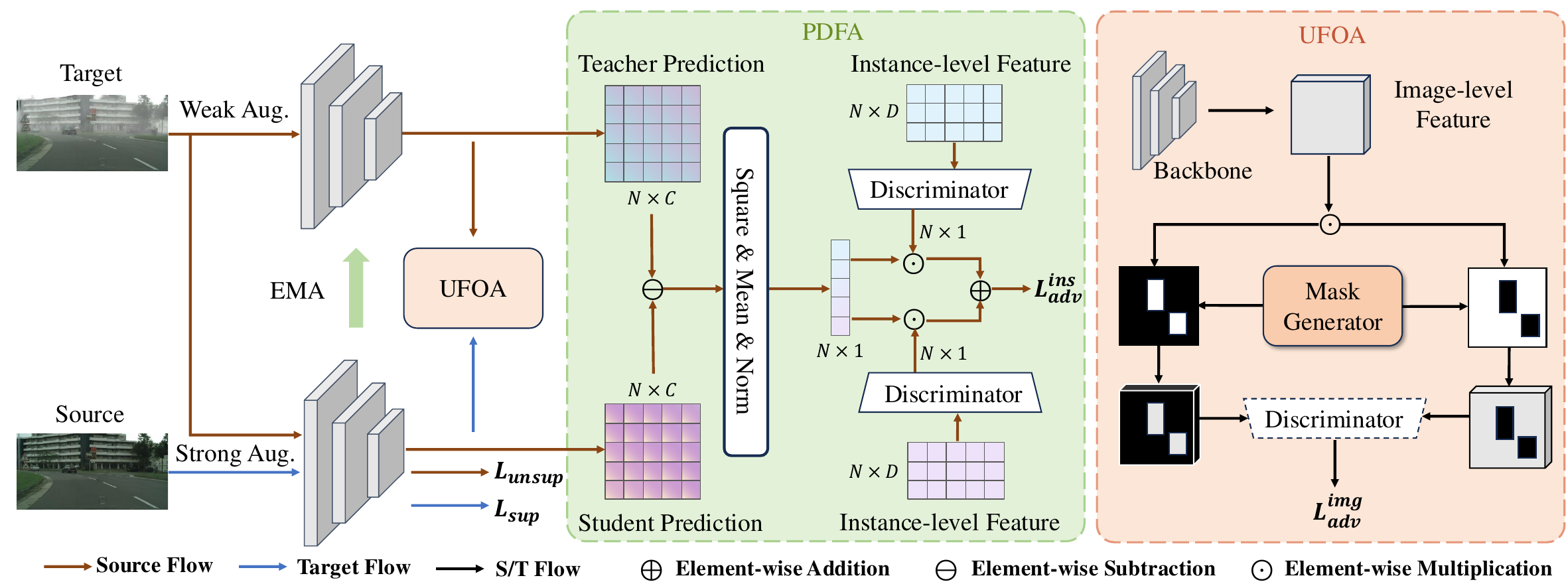}
    \caption{\textbf{Overview of our method.} Our approach is built upon the adaptive teacher-student framework. PDFA adjusts weights to different instances with respect to the discrepancy between predictions of the teacher and the student, while UFOA consists of a mask generator and an image-level discriminator. The mask generator produces a foreground-indicating mask to roughly separate the features of the last stage of the FPN into foreground and background parts.}
    \vspace*{-3mm}
    \label{fig:network}
\end{figure*}

\textbf{Domain Alignment via Adversarial Learning.} 
Domain alignment approaches advocate the use of adversarial learning to align distributions across domains.
As two classic works, DA-Faster~\cite{4Da-faster} and SADA~\cite{7SADA} propose image-level and instance-level alignment modules to alleviate the domain discrepancy. In addition to strong data augmentation, AT~\cite{11AT} also leverages image-level domain adaptive learning. 
MGA~\cite{64mga} introduces a unified multi-granularity alignment-based detection framework to learn domain-invariant representations and explore the relationship between features of different granularities in alignment. 
More recently, REACT~\cite{63react} adaptively compensates the extracted features with the remainder features for generating task-relevant features.
Among Def DETR-based~\cite{56def_detr} attempts, SFA~\cite{13sfa}, O$^2$net~\cite{15o2net} and MRT~\cite{17mrt} adopt domain query-based feature alignment, which is specially designed for domain adaptation of detection transformers. 
O$^2$net introduces object-aware alignment to emphasize foreground regions containing objects. Similar but different, our proposed strategy takes into account not only foreground but also background information with varying weights, as the alignment of background information is also indispensable. 

Motivated by the above, this work designs a differential feature alignment strategy to adjust the treatment of features in light of the varying importance of different regions.

\section{Methodology}
\subsection{Schematic Overview}
In the context of DAOD, suppose we have a set of $N_s$ images with category and object bounding-box labels, \emph{i.e.}, $\mathcal{D}_s=\{(\mathbf{X}_1, \mathbf{L}_1),..., (\mathbf{X}_{N_s}, \mathbf{L}_{N_s})\}$, from the source domain, and another set of $N_t$ unlabeled images, say $\mathcal{D}_t=\{\mathbf{Y}_1,...,\mathbf{Y}_{N_t}\}$, from the target domain. The goal of DAOD is to enhance the performance of detection on the target domain by leveraging labeled $\mathcal{D}_s$ and unlabeled $\mathcal{D}_t$. Due to the rationality, the adaptive teacher-student paradigm~\cite{4Da-faster} with several successful follow-ups~\cite{63react, 17mrt} has become popular in the field, which mainly adopts the self-training framework in conjunction with adversarial learning. Our proposed method also follows this technical route, as schematically illustrated in Fig.~\ref{fig:network}. Before launching our contributions, the core components of the paradigm shall be briefed for clarity.

\emph{Self-training framework}. As can be seen from Fig.~\ref{fig:network}, the teacher $\mathcal{T}$ and student networks $\mathcal{S}$ share the backbone and detector structures. The teacher processes each weakly augmented target image to generate pseudo-labels, while the student is optimized based on the supervision by the ground-truth labels of source-domain data and the pseudo labels of target-domain samples. The teacher is then updated by Exponential Moving Average (EMA)~\cite{19ema} from the student in the following manner: 
\begin{equation}
    \Theta_{\mathcal{T}}\leftarrow\alpha\Theta_{\mathcal{T}}+(1-\alpha)\Theta_{\mathcal{S}},
\end{equation}
where $\Theta_{\mathcal{T}}$ and $\Theta_{\mathcal{S}}$ denote the learnable parameters of $\mathcal{T}$ and $\mathcal{S}$, respectively. In addition, the hyper-parameter $\alpha\in[0,1]$ designates the smoothing factor of EMA. In our experiments, setting $\alpha$ to 0.9996 works sufficiently well.

\emph{Discriminators for adversarial alignment}. In addition to the teacher-student framework, domain discriminators are utilized to facilitate the alignment of feature distributions between the source and target domains. Specifically, the domain discriminators are strategically positioned after certain components to distinguish whether the feature is from the source domain or the target, which executes an adversarial learning process. 
In our implementation, we adopt Faster R-CNN as the detector, and set up two discriminators for the backbone and ROI head to assist with image-level and instance-level alignment, respectively. 

Although previous approaches have incorporated the aforementioned strategies, they usually overlook that different regions within an image contain varying amounts of domain-specific information. The failure to consider this aspect results in suboptimal alignment, particularly when domain-specific features differ across various spatial locations within an image. 
To address this limitation, this study builds two key modules to flexibly alter the alignment attention at two different levels, including an adaptive prediction discrepancy feedback instance-level alignment (PDFA), and an uncertainty-based foreground-guided image-level alignment (UFOA), as depicted in Fig.~\ref{fig:network}. The subsequent sections will detail these two designs.

\subsection{Prediction-Discrepancy Feedback Alignment}
\begin{figure}[!t]
    \centering
    \includegraphics[width=1.0\linewidth]{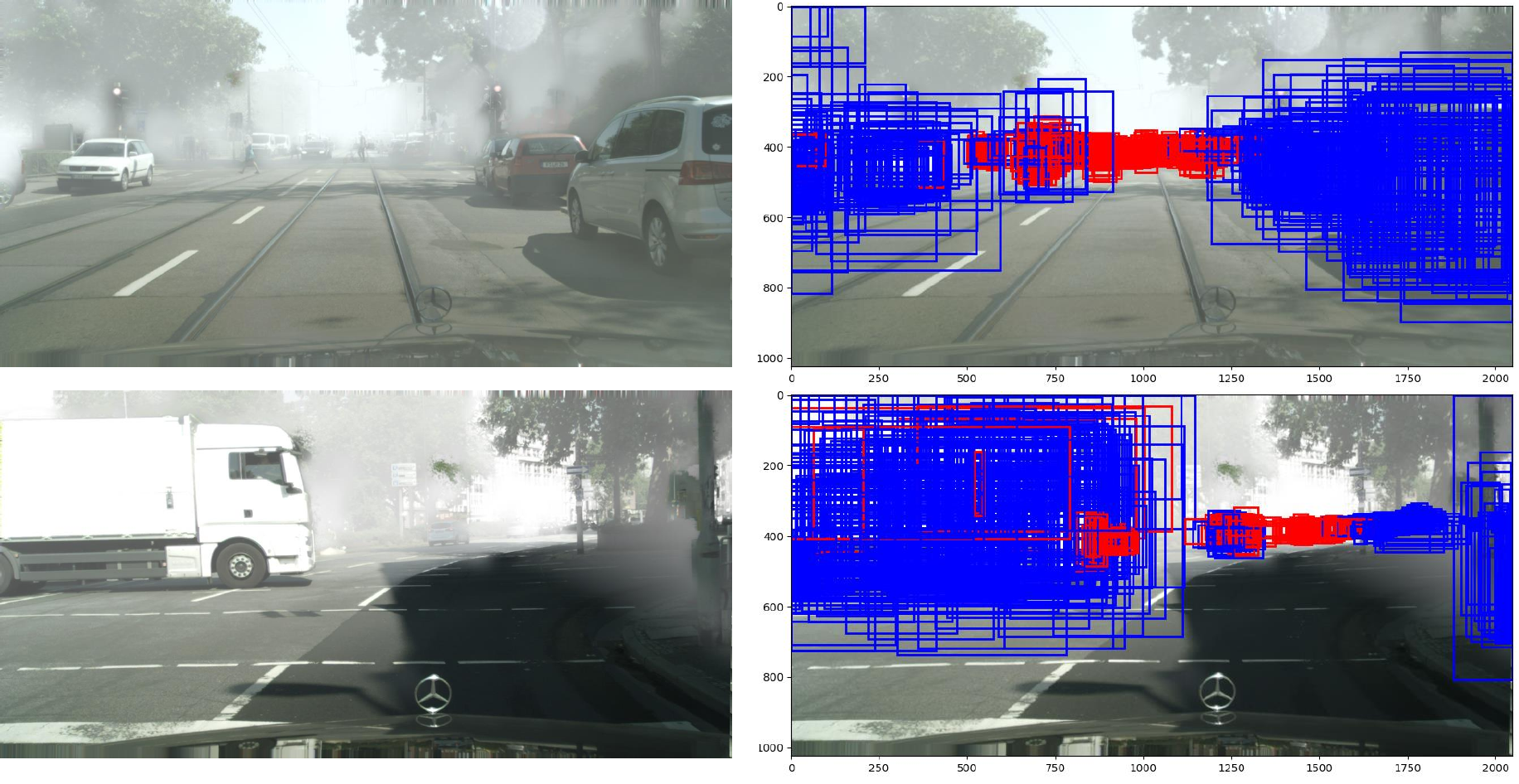}
    \caption{The proposals with top 2\% prediction discrepancies are marked in {\color{red}red}, while the rest are colored in {\color{blue}blue}.}
    \vspace*{-3mm}
    \label{fig:inconsistent-pred}
\end{figure}

Since the instance features generated by the ROI head of Faster R-CNN detector are the most proximal image features to final detection results, aligning these features seems to be direct and rational. Again, we emphasize that different instance regions may contain varying amounts of domain-specific information.
Figure~\ref{fig:inconsistent-pred} offers two examples, from which we can see that the fog density surrounding different cars in the images considerably differs. It is reasonable to deem that different instances should receive varying strengths of alignment.
To this end, a prediction-discrepancy feedback module is introduced to automatically identify how rich domain-specific information appears in a certain instance region, and determine how much alignment attention to pay on this instance accordingly.

\emph{Prediction-discrepancy feedback.} 
To automatically recognize instances with rich domain-specific information, we attempt to measure the prediction discrepancy between the teacher and student models on the same instances. Let us take a closer look at the right column of Fig.~\ref{fig:inconsistent-pred}. The red proposals stand for those of heavy prediction discrepancy between the teacher and student models, while the blue ones are of light discrepancy. As can be observed, those areas embracing more inconsistently predicted proposals tend to be of richer domain-specific information (denser fog in these cases). 
In the sequel, we guide the instance-level alignment by the measurement of instance-wise prediction discrepancy, making the model concentrate more on the alignment of regions with greater prediction discrepancies. Owing to the simplicity, this work directly employs the classification map as the prediction map. Given $N$ instance candidates and $C$ classes in total, the prediction discrepancy matrix $\textbf{P}_{\text{div}}\in\mathbb{R}^{N\times C}$ can be simply calculated as follows:
\begin{equation}
\mathbf{P}_{\text{div}}=\text{Square}(\mathbf{P}_{\mathcal{T}}-\mathbf{P}_{\mathcal{S}}),
\end{equation}
where $\mathbf{P}_{\mathcal{T}}\in \mathbb{R}^{N\times C}$ and $\mathbf{P}_{\mathcal{S}} \in \mathbb{R}^{N\times C}$ represent the prediction classification maps derived from the teacher and student models, respectively.

\emph{Weighted instance-level alignment.} Having $\mathbf{P}_{\text{div}}$ computed, we come to the construction of instance-wise alignment attention. The initial weight $\mathbf{w}_{\text{ins}}\in\mathbb{R}^{N\times1}$ can be simply obtained by:
\begin{equation}
    \mathbf{w}_{\text{ins}}=\frac1C\sum_{c}^C\mathbf{P}_{\text{div}}(:,c).
\end{equation}
By further applying the min-max normalization on $\mathbf{w}_{\text{ins}}$, the weight is restricted into the range of $[0,1]$ as follows: 
\begin{equation}
\mathbf{\tilde{w}}_{\text{ins}}=\frac{\mathbf{w}_{\text{ins}}- \min(\mathbf{w}_{\text{ins}})}{\max(\mathbf{w}_{\text{ins}}) - \min(\mathbf{w}_{\text{ins}})}.
\label{eq:wight}
\end{equation}
Moreover, the instance-wise adversarial loss is computed by: 
\begin{equation}
    \mathbf{f}_{\textbf{ins}}= -\mathbf{d}\odot\log(\mathfrak{D}_1(\mathbf{F}_{\text{ins}}))-\mathbf{\bar{d}}\odot\log(1- \mathfrak{D}_1(\mathbf{F}_{\text{ins}})),
    \label{eq:advins}
\end{equation} 
where $\odot$ means Hadamard product. Moreover, $\mathbf{d} \in \{0, 1\}^{N\times 1}$ is the domain flag vector, and $\mathbf{\bar{d}}$ is the complement version of $\mathbf{d}$. $\mathfrak{D}_1(\cdot)$ represents the instance discriminator whose input is the collection of $N$ instance candidates $\mathbf{F}_{\text{ins}} \in \mathbb{R}^{N\times D}$. Its outputs the domain discriminating results $\mathfrak{D}_1(\mathbf{F}_{\text{ins}}) \in \mathbb{R}^{N\times 1}$.
Combining Eqs.~\eqref{eq:wight} and~\eqref{eq:advins} yields final weighted instance-level adversarial loss:
\begin{equation}
    \mathcal{L}_{adv}^{ins} =  \|\mathbf{\tilde{w}}_{\text{ins}}\odot \mathbf{f}_{\text{ins}}\|_1,
\end{equation}
where $\|\cdot\|_1$ represents the $\ell_1$ norm.

\subsection{Uncertain-based Foreground-Oriented Alignment}
As previously discussed, another functionality is required to explicitly guide the model by prioritizing the alignment of regions containing foreground objects. This part details our foreground-oriented alignment scheme. It comprises a mask generator for (approximately) indicating foreground and background areas of an image, and splitting the feature maps into two parts accordingly for subsequent discrimination and reweighting operations. 

\begin{figure}[!t]
    \centering
    \includegraphics[width=1.0\linewidth]{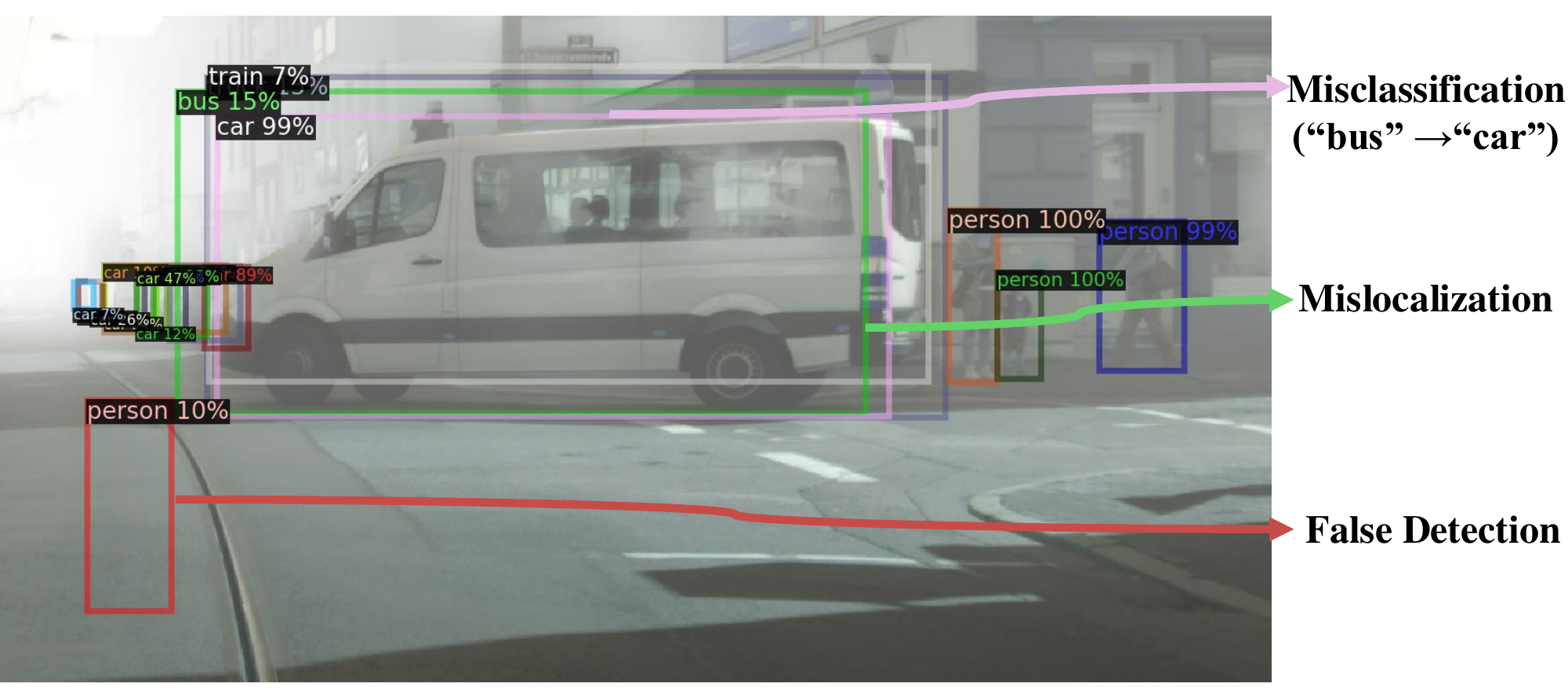}
    \caption{Visualization of pseudo labels generated by the teacher model. Despite misclassification, mislocalization and false detection errors exist, the union of these inaccurate bounding boxes can still largely indicate foreground areas.}
    \vspace*{-3mm}
    \label{fig:inaccu}
\end{figure}

\emph{Mask generator.} It is not difficult to divide images into foreground and background areas, if with the help of ground-truth annotations. However, for target-domain data, the ground truths are absent. Alternatively, we resort to the pseudo labels generated by the teacher model. Figure~\ref{fig:inaccu} exhibits an example marked with the generated pseudo labels. We can observe that, although these pseudo bounding boxes may be inaccurate, their union can largely zone foreground areas. Hence, we form the foreground mask $\mathbf{M}$ by utilizing the union of detection boxes: for the labeled source domain, ground-truth bounding boxes are used to construct the mask, \emph{i.e.} the elements within the regions enclosed by these bounding boxes are set to 1 (and 0 otherwise), while for the unlabeled target domain, detected bounding-boxes by the teacher model (pseudo labels) are employed to accomplish the mask. As a consequence, the image feature $\mathbf{F}_{\text{img}}$ can be split simply via:
\begin{equation}
\begin{aligned}
    \mathbf{F}^{fg}_{\text{img}} = \mathbf{M} \odot \mathbf{F}_{\text{img}},\ \
    \mathbf{F}^{bg}_{\text{img}} = \mathbf{\bar{M}} \odot \mathbf{F}_{\text{img}},
\end{aligned}
\end{equation}
where the feature F\_img refers to the P2 layer within the FPN, which is the highest-resolution feature map in the FPN, capturing fine-grained spatial details. $\mathbf{F}^{fg}_{\text{img}}$ and $\mathbf{F}^{bg}_{\text{img}}$ correspond to foreground and background feature parts, respectively. In addition, $\mathbf{\bar{M}}$ represents the complementary of $\mathbf{M}$.

\emph{Uncertainty-based image-level alignment.} It is worth noting that, different from O$^2$net~\cite{15o2net}, our UFOA module retains background regions along with regions of interest, which are both fed into the discriminator. 
This operation ensures that while foreground information remains the primary focus during alignment, background information is also considered, preventing it from being entirely discarded. Subsequently, an uncertainty factor is introduced to balance the relative importance of these two types of information during image-level alignment.
The respective adversarial losses on the separated regions of interest and background regions are obtained through:
\begin{equation}
\begin{split}
    \mathcal{L}_{adv}^{fg}= -d\log(\mathfrak{D}_2(\mathbf{F}_{\text{img}}^{fg}))-\bar{d}\log(1- \mathfrak{D}_2(\mathbf{F}_{\text{img}}^{fg})),\\
    \mathcal{L}_{adv}^{bg}= -d\log(\mathfrak{D}_2(\mathbf{F}_{\text{img}}^{bg}))-\bar{d}\log(1- \mathfrak{D}_2(\mathbf{F}_{\text{img}}^{bg})).
\end{split}
\end{equation}
Note that $\mathfrak{D}_2(\cdot)$ is the image-level discriminator, which takes the feature $\mathbf{F}_{\text{img}}$ as input and outputs the domain discriminating result $ \mathfrak{D}_2(\mathbf{F}_{\text{img}}^{f}) \in [0, 1]$ for the entire image-level feature. Here, $d \in \{0, 1\}$ is a domain flag. Finally, the adversarial loss at the image level is a weighted sum of the above two terms as:
\begin{equation}
    \mathcal{L}_{adv}^{img} = \gamma \mathcal{L}_{adv}^{fg} + (1 - \gamma) \mathcal{L}_{adv}^{bg}.
    \label{eq:balance}
\end{equation}
By tuning the hyper-parameter uncertainty factor $\gamma$, we can modulate the relative emphasis on aligning these two components at the image level, thereby facilitating differential alignment. Notably, when setting $\gamma=1$ in Eq.~\eqref{eq:balance}, only foreground regions are taken into account in the image-level alignment, as the alignment pattern in~\cite{15o2net}. By contrast, our alignment pattern is a balanced foreground-background alignment, the superiority of which will be validated in the ablation study.

\subsection{Overall Objective Function}
Combining all the presented parts, the overall objective function turns out to be:
\begin{equation}
\max_{\mathfrak{D}_1,\mathfrak{D}_2}\min_{\mathfrak{G}}\mathcal{L}_{sup}+\mathcal{L}_{unsup}{+\lambda}(\mathcal{L}_{adv}^{ins}+\mathcal{L}_{adv}^{img}),
\end{equation}
where $\mathfrak{G}$ designates the feature extractor in the network. It includes all learnable components except for the two discriminators.
In our experiments, we set $\lambda$ as 0.01 by default. To be clear, the supervised detection loss $\mathcal{L}_{sup}$ is calculated with respect to ground truth labels, while $\mathcal{L}_{unsup}$ receives supervision from pseudo labels as: 
\begin{equation}
\begin{aligned}
    \mathcal{L}_{sup} = \mathcal{L}^{s}_{cls}(\Tilde{c}_s, c_s) + \mathcal{L}^{s}_{reg}(\Tilde{b}_s, b_s), \\ 
    \mathcal{L}_{unsup} = \mathcal{L}^{t}_{cls}(\Tilde{c}_t, \hat{c}_t) + \mathcal{L}^{t}_{reg}(\Tilde{b}_t, \hat{b}_t),
\end{aligned}
\end{equation}
where $c$, $\Tilde{c}$, and $\hat{c}$ stand for the classification ground truth of source domain, the classification result and pseudo-classification label, respectively. The same principle applies to the bounding-box symbols $b$, $\Tilde{b}$, and $\hat{b}$. $\mathcal{L}_{cls}$ adopts the cross-entropy loss for classification, and $\mathcal{L}_{reg}$ is the $\ell_1$ loss used for regression.

\begin{table*}[!ht]
\centering
\resizebox{\textwidth}{!}{
\begin{tabular*}{\textwidth}{llccccccccc}
    \toprule
    \textbf{Method} & \textbf{Detector} & \textbf{person} & \textbf{rider} & \textbf{car} & \textbf{truck} & \textbf{bus} & \textbf{train} & \textbf{mcycle} & \textbf{bicycle} & \textbf{AP$^{val}_{50}$} \\ 
    \midrule
    FCOS~\cite{66fcos} & FCOS & 29.6 & 26.3 & 37.1 & 7.9 & 14.1 & 6.3 & 12.9 & 28.1 & 20.3 \\
    EPM~\cite{59epm} & FCOS & 41.9 & 38.7 & 56.7 & 22.6 & 41.5 & 26.8 & 24.6 & 35.5 & 36.0 \\
    SIGMA~\cite{10sigma} & FCOS & 44.0 & 43.9 & 60.3 & 31.6 & 50.4 & 51.5 & 31.7 & 40.6 & 44.2 \\
    CSDA~\cite{65csda} & FCOS & 46.6 & 46.3 & 63.1 & 28.1 & 56.3 & 53.7 & 33.1 & 39.1 & 45.8 \\
    HT~\cite{62ht} & FCOS & 52.1 & 55.8 & 67.5 & 32.7 & 55.9 & 49.1 & 40.1 & 50.3 & 50.4 \\    
    \midrule

    Def DETR~\cite{56def_detr} & Def DETR & 37.7 & 39.1 & 44.2 & 17.2 & 26.8 & 5.8 & 21.6 & 35.5 & 28.5 \\ 
    SFA~\cite{13sfa} & Def DETR & 46.5 & 48.6 & 62.6 & 25.1 & 46.2 & 29.4 & 28.3 & 44.0 & 41.3 \\ 
    O$^2$net~\cite{15o2net} & Def DETR & 48.7 & 51.5 & 63.6 & 31.1 & 47.6 & 47.8 & 38.0 & 45.9 & 46.8 \\ 
    MRT~\cite{17mrt} & Def DETR & \underline{52.8} & 51.7 & \underline{68.7} & \underline{35.9} & \underline{58.1}& \underline{54.5} & 41.0 & 47.1 & 51.2 \\
    \midrule

    Faster R-CNN~\cite{45fastrcnn} & FRCNN & 26.9 & 38.2 & 35.6 & 18.3 & 32.4 & 9.6 & 25.8 & 28.6 & 26.9 \\
    DA-Faster~\cite{4Da-faster} & FRCNN & 29.2 & 40.4 & 43.4 & 19.7 & 38.3 & 28.5 & 23.7 & 32.7 & 32.0 \\
    SADA~\cite{7SADA} & FRCNN & 48.5 & 52.6 & 62.1 & 29.5 & 50.3 & 31.5 & 32.4 & 45.4 & 44.0 \\
    PT~\cite{3PT} & FRCNN & 40.2 & 48.8 & 63.4 & 30.7 & 51.8 & 30.6 & 35.4 & 44.5 & 42.7 \\
    AT$^*$~\cite{11AT} & FRCNN & 43.7 & 54.1 & 62.3 & 31.9 & 54.4 & 49.3 & 35.2 & \underline{47.9} & 47.4 \\ 
    MIC~\cite{12mic} & FRCNN & 50.9 & 55.3 & 67.0 & 33.9 & 52.4 & 33.7 & 40.6 & 47.5 & 47.6 \\
    MGA~\cite{64mga} & FRCNN & 47.0 & 54.6 & 64.8 & 28.5 & 52.1 & 41.5 & 40.9 & 49.5 & 47.4 \\
    REACT~\cite{63react} & FRCNN & 52.1 & \underline{57.1} & 66.3 & 35.0 & 56.7 & 52.8 & \underline{42.9} & \underline{53.8} & \underline{52.1} \\
    \midrule
    
    Ours & FRCNN & \textbf{59.8} & \textbf{62.8} & \textbf{73.7} & \textbf{40.3} & \textbf{59.4} & \textbf{56.1} & \textbf{47.8} & \textbf{58.3} & \textbf{57.3} \\
    \bottomrule
\end{tabular*}
}
\vspace*{-2mm}
\caption{
Results on adaptation from Cityscapes to Foggy Cityscapes (0.02). AT$^*$ denotes that the results of AT on \textit{Foggy (0.02)} are taken from~\cite{17mrt}. The best results are in \textbf{bold}, while the second-best results are \underline{underlined}.}
\vspace*{-4mm}
\label{tab:ct}
\end{table*}

\begin{table*}[t]
\centering
\begin{tabular*}{\textwidth}{llc@{\hspace{16pt}}c@{\hspace{16pt}}c@{\hspace{16pt}}c@{\hspace{16pt}}c@{\hspace{16pt}}c@{\hspace{16pt}}c@{\hspace{16pt}}c}
    \toprule
          \textbf{Method} & \textbf{Detector} & \textbf{person} & \textbf{rider} & \textbf{car} & \textbf{truck} & \textbf{bus} & \textbf{mcycle} & \textbf{bicycle} & \textbf{AP$^{val}_{50}$} \\ 
        \midrule
        EPM~\cite{59epm} & FCOS & 39.6 & 26.8 & 55.8 & 18.8 & 19.1 & 14.5 & 20.1 & 27.8 \\
        SIGMA~\cite{10sigma} & FCOS & 46.9 & 29.6 & \underline{64.1} & 20.2 & 23.6 & 17.9 & 26.3 & 32.7 \\
        HT~\cite{62ht} & FCOS & \underline{53.4} & \underline{40.4} & 63.5 & \underline{27.4} & \underline{30.6} & \underline{28.2} & \textbf{38.0} & \underline{40.2} \\
        
        \midrule
        Def DETR~\cite{56def_detr} & Def DETR & 38.9 & 26.7 & 55.2 & 15.7 & 19.7 & 10.8 & 16.2 & 26.2 \\ 
        SFA~\cite{13sfa} & Def DETR & 40.2 & 27.6 & 57.5 & 19.1 & 23.4 & 15.4 & 19.2 & 28.9 \\ 
        AQT~\cite{16aqt} & Def DETR & 38.2 & 33.0 & 58.4 & 17.3 & 18.4 & 16.9 & 23.5 & 29.4 \\ 
        O$^2$net~\cite{15o2net} & Def DETR & 40.4 & 31.2 & 58.6 & 20.4 & 25.0 & 14.9 & 22.7 & 30.5 \\ 
        MTTrans~\cite{14mttrans} & Def DETR & 44.1 & 30.1 & 61.5 & 25.1 & 26.9 & 17.7 & 23.0 & 32.6 \\ 
        MRT~\cite{17mrt} & Def DETR & 48.4 & 30.9 & 63.7 & 24.7 & 25.5 & 20.2 & 22.6 & 33.7 \\ 

        \midrule
        Faster R-CNN~\cite{45fastrcnn} & FRCNN & 28.8 & 25.4 & 44.1 & 17.9 & 16.1 & 13.9 & 22.4 & 24.1 \\
        DA-Faster~\cite{4Da-faster} & FRCNN & 28.9 & 27.4 & 44.2 & 19.1 & 18.0 & 14.2 & 22.4 & 24.9 \\
        ICR-CCR-SW~\cite{58icr} & FRCNN & 32.8 & 29.3 & 45.8 & 22.7 & 20.6 & 14.9 & 25.5 & 27.4 \\ 
        REACT~\cite{63react} & FRCNN & - & - & - & - & - & - & - & 35.8 \\

        \midrule
        Ours & FRCNN & \textbf{61.4} & \textbf{45.4} & \textbf{75.4} & \textbf{33.0} & \textbf{36.2} & \textbf{29.5} & \underline{36.7} & \textbf{45.8} \\ 
    \bottomrule[1pt]
\end{tabular*}
\vspace*{-1mm}
\caption{Results on adaptation from Cityscapes to BDD100k-daytime. The best results are in \textbf{bold}, while the second-best results are \underline{underlined}.} 
\vspace*{-3mm}
\label{tab:bdd100k}
\end{table*}

\begin{table}[!t]
\centering
\begin{tabular}{l@{\hspace{8.0pt}}c@{\hspace{8.0pt}}c}
    \toprule
       \textbf{Method} & \textbf{Detector} & \textbf{carAP$^{val}_{50}$} \\ 
       \midrule
        FCOS~\cite{66fcos} & FCOS & 39.8 \\
        EPM~\cite{59epm} & FCOS & 49.0 \\
        SIGMA~\cite{10sigma} & FCOS & 53.7 \\
        HT~\cite{62ht} & FCOS & \underline{65.5} \\ 
        
       \midrule
        Def DETR~\cite{56def_detr}& Def DETR & 47.4 \\ 
        SFA~\cite{13sfa} & Def DETR & 52.6 \\ 
        O$^2$net~\cite{15o2net} & Def DETR & 54.1 \\ 
        MTTrans~\cite{14mttrans} & Def DETR & 57.9 \\ 
        MRT~\cite{17mrt} & Def DETR & 62.0 \\ 

        \midrule
        Faster R-CNN~\cite{1fasterrcnn} & FRCNN & 39.4 \\ 
        DA-Faster~\cite{4Da-faster} & FRCNN & 41.9 \\ 
        MeGA-CDA~\cite{40mega} & FRCNN & 44.8 \\ 
        D-adapt~\cite{9D-adapt} & FRCNN & 51.9 \\
        PT~\cite{3PT} & FRCNN & 55.1 \\ 
        REACT~\cite{63react} & FRCNN & 58.6 \\

        \midrule
        Ours & FRCNN & \textbf{69.7} \\
    \bottomrule[1pt]
\end{tabular}
\vspace*{-1mm}
\caption{Results on adaptation from Sim10k to Cityscapes (category `car'). }
\vspace*{-6mm}
\label{tab:sim10k}
\end{table}

\section{Experimental Validation}
\subsection{Datasets} Following recent DAOD approaches~\cite{62ht, 17mrt, 63react}, we evaluate our method on three widely-used benchmarks. Specifically, we perform adaptation experiments on three common scenarios: (1) weather adaptation with Cityscapes $\rightarrow$ Foggy Cityscapes, (2) synthetic to real adaptation with Sim10k $\rightarrow$ Cityscapes, and (3) small to large-scale dataset adaptation with Cityscapes $\rightarrow$ BDD100K-daytime. 

\textbf{Cityscapes}~\cite{23cityscapes} comprises 2,975 training images and 500 validation images, covering various urban environments and traffic conditions. For our experiments, the semantic segmentation labels provided by Cityscapes are converted into bounding box annotations, allowing the dataset to be repurposed for object detection tasks.

\textbf{Foggy Cityscapes}~\cite{38syn-foggy} is a synthetic dataset rendered from Cityscapes with three levels of foggy density (0.005, 0.01, 0.02). We use the highest density (0.02) as the target domain for a fair comparison following existing methods~\cite{63react,64mga,17mrt,15o2net,13sfa}. 

\textbf{Sim10k}~\cite{51sim10k} contains 10,000 images rendered from GTA engine. In one of our adaptation experiments, Sim10k is used as the source domain, while the Cityscapes dataset, representing real-world scenes, serves as the target domain. The experiment focuses specially on the detection of `car' category, aiming to evaluate the model's ability to adapt from synthetic to real-world scenarios. 

\textbf{BDD100k-daytime}~\cite{52bdd100k} is a subset of the larger BDD100k datset, designed for autonomous driving research and various computer vision tasks. It contains 36,728 training images and 5,258 validation images, which provides a diverse environment that mirrors real-world daytime driving, making it crucial for evaluating and improving detection models across domains. 

\subsection{Implementation Details} 
\textbf{Network architecture.}
We use Faster R-CNN~\cite{45fastrcnn} as our base detector, with ResNet-50~\cite{53resnet} pre-trained on ImageNet~\cite{54imagenet} and Feature Pyramid Network~\cite{34fpn} as the backbone. The teacher model is updated only by EMA from the student model, and we freeze the first two layers of the ResNet backbone network. Our model is implemented in PyTorch and trained on 4 NVIDIA RTX3090 GPUs with 24 GB of memory each. Our implementation is built upon Detectron2 with default settings including multi-scale input transforms.

\textbf{Data augmentations.} We employ common weak augmentation techniques in DAOD, including RandomContrast, RandomBrightness, RandomSaturation, RandomGrayscale, and RandomBlur. For the source images, strong augmentation consists of the weak augmentations combined with the Cutout~\cite{24cutout} method, while for target domain images, it includes the weak augmentation along with the MIC~\cite{12mic} method. 

\textbf{Optimization.} We optimize the network using the SGD optimizer with a momentum of 0.9. The initial learning rate is set to 0.01 and decreases after 1600 iterations. We use a batch size of 32, consisting of 16 labeled source images and 16 unlabeled target ones, and train the network for 25k iterations in total, including 10,000 iterations for burn-in and 15,000 iterations for teacher-student mutual learning. 

\textbf{Evaluation metric.} We assess adaptation performance by reporting mean Average Precision (mAP) with an IoU threshold of 0.5, following standard protocols on the three aforementioned benchmarks. The results for prior methods are based on the values reported in their original papers.

\subsection{Comparisons with Other Methods}
\textbf{Adaptation from normal to foggy weather.}
The adaptation results of Cityscapes to Foggy Cityscapes are shown in Table~\ref{tab:ct}. 
Ours achieves the best AP$_{50}$ of 57.3\% and surpasses the second best REACT~\cite{63react} with 52.1\% by a margin of 5.2\%. Notably, our method demonstrates a marked improvement in the first three categories: person (+7.7\%), rider (+5.7\%), and car (+7.4\%). The significant performance gain in the long-tail category `rider' further underscores the efficacy of our differential alignment strategy, which enables the model to better align regions that have a greater impact on detection results. 

\textbf{Adaptation from small to large-scale dataset.}
Table~\ref{tab:bdd100k} presents the adaptation performance from Cityscapes to BDD100K. Our method achieves 45.8\% AP$_{50}$, outperforming all baseline models by a considerable margin. Specifically, it outperforms the best performing one-stage adaptive detector, HT, by 5.6\% AP$_{50}$, and exceeds the two-stage DAOD approach REACT~\cite{63react}, by 10.0\% AP$_{50}$.
Although our method does not surpass the current state-of-the-art in the `bicycle' category, it achieves the second-best performance, demonstrating competitive effectiveness. Despite this, our method shows superior performance in most categories, underscoring its robustness and generalizability.

\textbf{Adaptation from synthetic to real images.}
We investigate the synthetic-to-real domain adaptation scenario, specifically evaluating the adaptation from Sim10K to Cityscapes, with results presented in Tab.~\ref{tab:sim10k}. Our method achieves 69.7\% AP$_{50}$, surpassing previous state-of-the-art method by a margin of 4.2\%. Sim-to-Real adaptation task requires transferring the semantics of a synthetic scene from the GTA engine to a real scene, and our performance improvement on this task also demonstrates robustness in the face of synthetic scenarios.

\subsection{Ablation Studies}
\textbf{Ablation on network components.}
We conduct ablation studies to evaluate the significance of key network components, as shown in Tab.~\ref{tab:ablation}. We replace PDFA and UFOA with standard instance-level and image-level alignment modules, which employ an equal alignment strategy. The replacements lead to 2.1\% and 1.1\% decreases in AP$_{50}$, respectively, underscoring the effectiveness of our proposed differential alignment strategy. 
Additionally, we observe that MIC strong augmentation significantly improve the model's perception of target domain distributions. 
Specifically, the baseline consists of a teacher-student framework, image-level and instance-level alignment modules, without the differential attention mechanisms introduced by PDFA and UFOA, as well as excluding the burn-in phase and mutual learning training process.

\textbf{Ablation on uncertainty factor \textbf{$\gamma$}.}
The results of different $\gamma$ in Eq.~\eqref{eq:balance} is shown in Tab.~\ref{tab:gamma ablation}, which highlights the effectiveness of our foreground-oriented alignment module. When the value of $\gamma$ is greater than 0.5, i.e. more attention is given to the foreground regions, the model obtains a greater performance improvement. But when the value of $\gamma$ is equal to 1.0 (\emph{i.e.} foreground alignment pattern in O$^2$net), the performance decreases due to the lack of alignment of the background information, validating the necessity of our balanced foreground-background alignment pattern.

\begin{table}[!t]
\centering
\begin{tabular}{c@{\hspace{15.0pt}}c@{\hspace{15.0pt}}c@{\hspace{15.0pt}}c@{\hspace{15.0pt}}c}
    \toprule
       \textbf{Baseline} & \textbf{Strong Aug} & \textbf{PDFA} & \textbf{UFOA} &  \textbf{AP$_{50}^{val}$} \\
       \midrule
       \checkmark &  &  &  & 50.0 \\ 
        \checkmark & \checkmark &  &  & 53.9 \\ 
        \checkmark & \checkmark & \checkmark &  & 56.2 \\ 
        \checkmark & \checkmark & & \checkmark & 55.2 \\
        \checkmark & \checkmark & \checkmark & \checkmark & 57.3 \\
    \bottomrule[1pt]
\end{tabular}
\vspace*{-1mm}
\caption{Ablation of proposed modules on adaptation from Cityscapes to Foggy Cityscapes.}
\vspace*{-3mm}
\label{tab:ablation}
\end{table}

\begin{table}[!t]
\centering
\begin{tabular}{c|@{\hspace{10.0pt}}c@{\hspace{38.0pt}}c@{\hspace{38.0pt}}c@{\hspace{38.0pt}}c}
    \toprule
       $\gamma$ & 0 & 0.5 & 0.8 & 1.0\\
       \midrule
       AP$_{50}^{val}$ & 55.1 & 55.9 & 57.3 & 55.6 \\
    \bottomrule[1pt]
\end{tabular}
\vspace*{-1mm}
\caption{Effect of the hyper-paramter $\gamma$ in UFOA on adaptation from Cityscapes to Foggy Cityscapes.}
\vspace*{-5mm}
\label{tab:gamma ablation}
\end{table}

\section{Conclusion}
In this paper, we tackled the issue of ineffective alignment in domain adaptive object detection by introducing two innovative modules: the adaptive Prediction-Discrepancy Feedback instance Alignment (dubbed PDFA) and the Uncertainty-based Foreground-Oriented image Alignment (UFOA). The PDFA module prioritizes instances with higher teacher-student prediction discrepancies, ensuring more accurate alignment of critical domain-specific features. Furthermore, the UFOA module guides the model's attention toward foreground regions, effectively mitigating the limitations of previous equal alignment strategies.
Comprehensive evaluations on widely used DAOD datasets, along with ablation studies, have validated the effectiveness of our proposed method and demonstrated its significant advantages over other state-of-the-art approaches.

\appendix
\section{Acknowledgments}
This work was supported by the National Natural Science Foundation of China under Grant numbers 62372251 and 62072327.

\bigskip
\bibliography{aaai25}

\clearpage

\twocolumn[
\section*{Supplementary Material for ``Differential Alignment for Domain Adaptive Object Detection"}
]

\noindent 
In the supplementary material, we provide detailed implementation specifics for PDFA, particularly addressing how we compute the prediction discrepancy between the teacher and student models, even when they may generate different sets of proposals. Additionally, we offer a further explanation of the experimental setup for the hyper-parameter $\gamma$ ablation study. We also present visualization results, including feature distribution visualizations and detection bounding box visualizations. Finally, we analyze the limitation of our method by presenting some failure cases observed in the adaptation from Sim10k~\cite{51sim10k} to Cityscapes~\cite{23cityscapes}.

\vspace{3mm}
\section{Implementation Details for PDFA}
In this section, we provide additional implementation details of the Prediction-Discrepancy Feedback instance Alignment (PDFA), which aims to adaptively assign higher weights to instances with higher teacher-student detection discrepancies, effectively handling heavier domain-specific information. The PDFA module calculates the discrepancy between the instance classification prediction maps generated by the teacher and student models. This process requires the proposal boxes generated by both the teacher and student models to be consistent; however, this is often not the case. To address this, in our implementation, we replace teacher proposal boxes with those generated by the student model, allowing for accurate calculation of instance-wise prediction discrepancies.

\section{Explanation of $\gamma$ Ablation Experimental Setup}
In the ablation study, we evaluate the effect of the hyper-parameter $\gamma$ in UFOA on adaptation from Cityscapes~\cite{23cityscapes} to Foggy Cityscapes~\cite{38syn-foggy}. Specifically, we access the model's performance with $\gamma$ set to \{ 0, 0.5, 0.8, 1.0\}, and confirm that the performance at $\gamma = 0.8$ (as set in our experiments) surpasses that at $\gamma = 1.0$. It is important to note that $\gamma = 0.8$ is not necessarily the optimal value; the optimal value may lie within [0.5, 0.8] or [0.8, 1.0], assuming that the entire curve follows a parabolic shape. Moreover, it is crucial to emphasize that the purpose of this experiment is not to precisely determine the optimal value of $\gamma$, but to demonstrate that (1) the optimal value of $\gamma$ lies within the range [0.5, 1.0), which validates the effectiveness of our foreground-oriented alignment pattern; and (2) a balanced foreground-background alignment pattern ($\gamma \neq 1.0$) is superior to a pattern that aligns only foreground information ($\gamma = 1.0$). Our experimental results effectively support these two points.

\begin{figure}[!t]
    \centering
    \vspace{4mm}
    \includegraphics[width=1.0\linewidth]{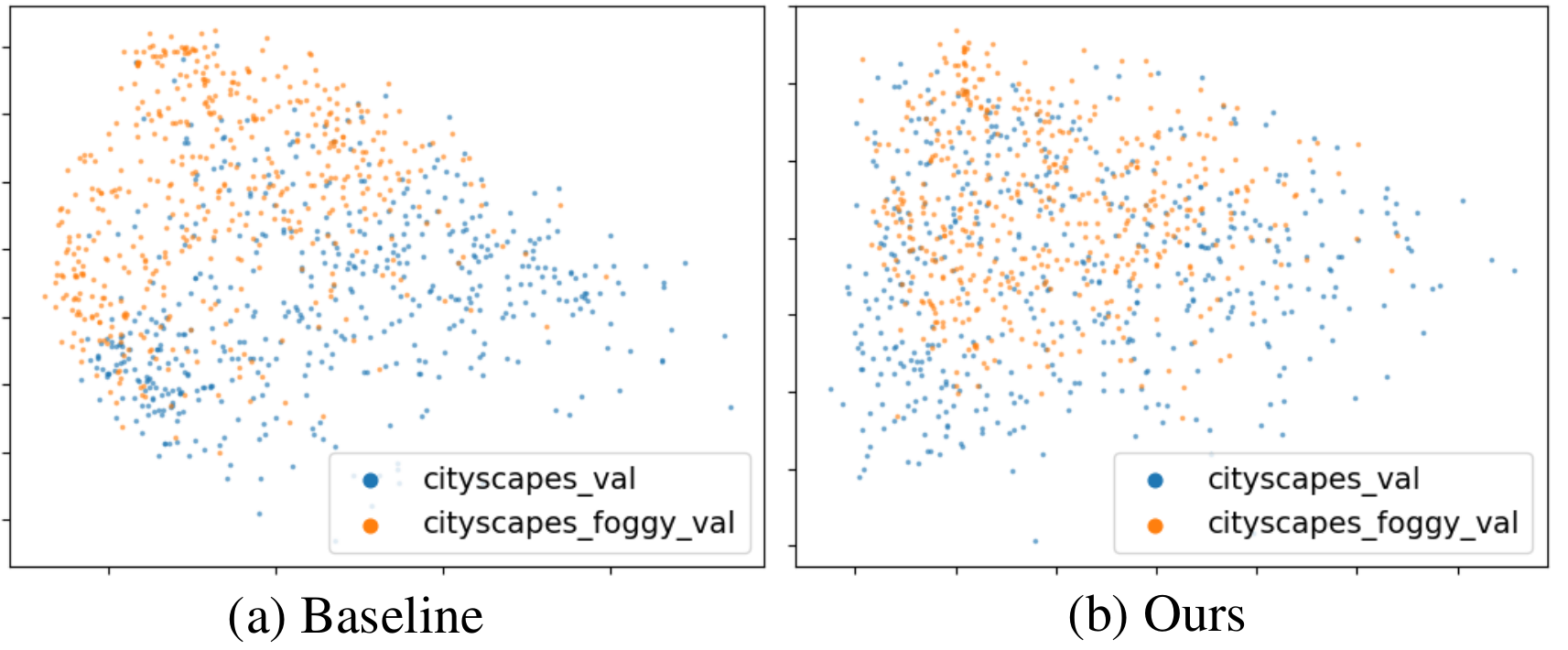}
    \caption{Feature distribution visualizations using PCA. Different colors represent different domains.}
    \vspace{-2mm}
    \label{fig:pca-vis}
\end{figure}

\section{Qualitative Results}
\subsection{Feature Distribution Visualizations}
We use PCA~\cite{69pca} to visualize the features extracted by the backbone. As shown in Figure~\ref{fig:pca-vis}, the features extracted by the baseline model can be easily separated by domain. In contrast, our proposed method learns domain-invariant features, making it difficult to distinguish between the features of the two domains.

\subsection{Comparisons with Baseline Model}
\begin{figure*}[htbp]
    \centering
    \includegraphics[width=1.0\linewidth]{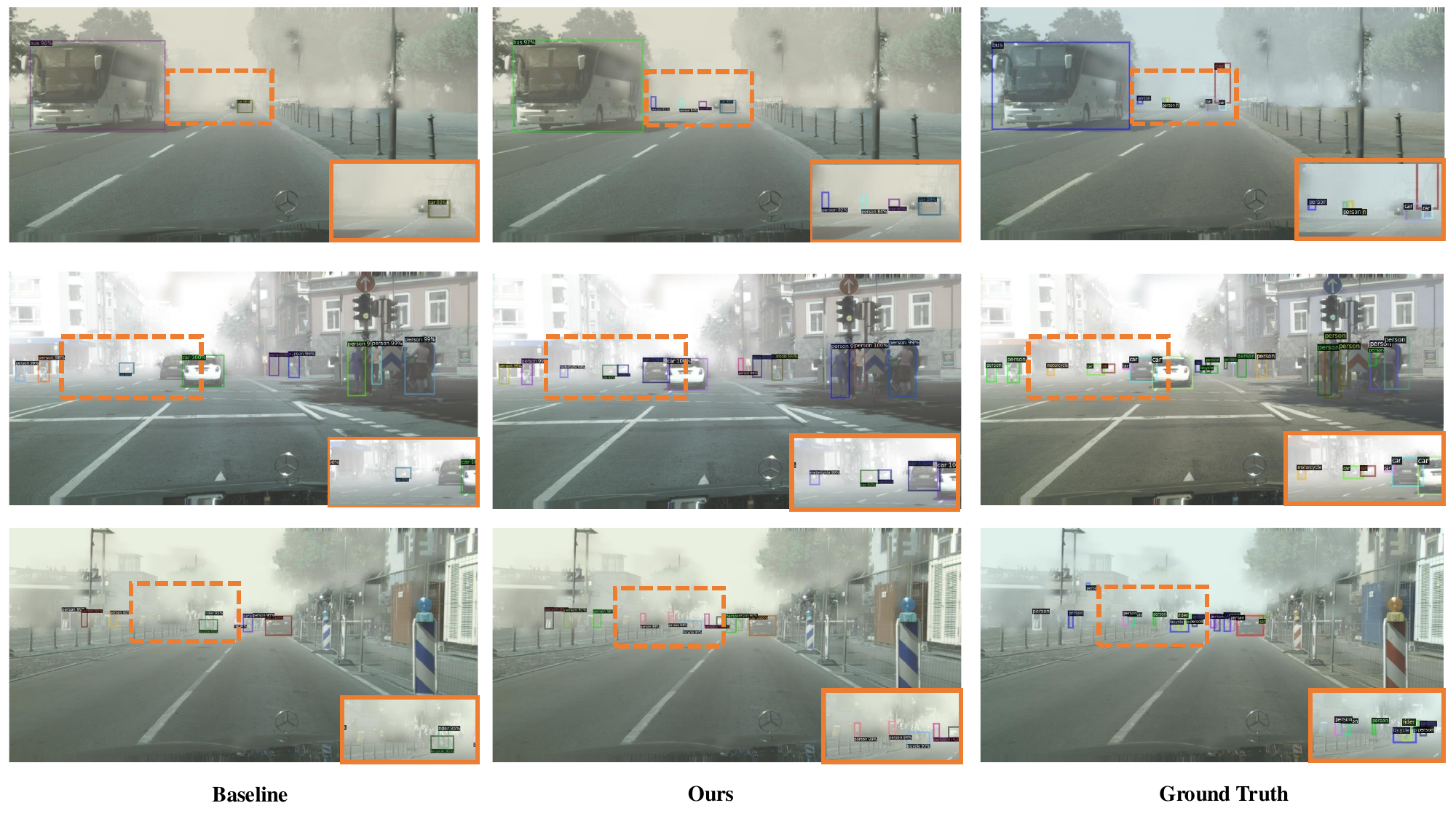}
    \caption{Qualitative results on adaptation from Cityscapes to Foggy Cityscapes. Please zoom in for details.}
    \vspace{-2mm}
    \label{fig:vis-c2f}
\end{figure*}

\begin{figure*}[htbp]
    \centering
    \includegraphics[width=1.0\linewidth]{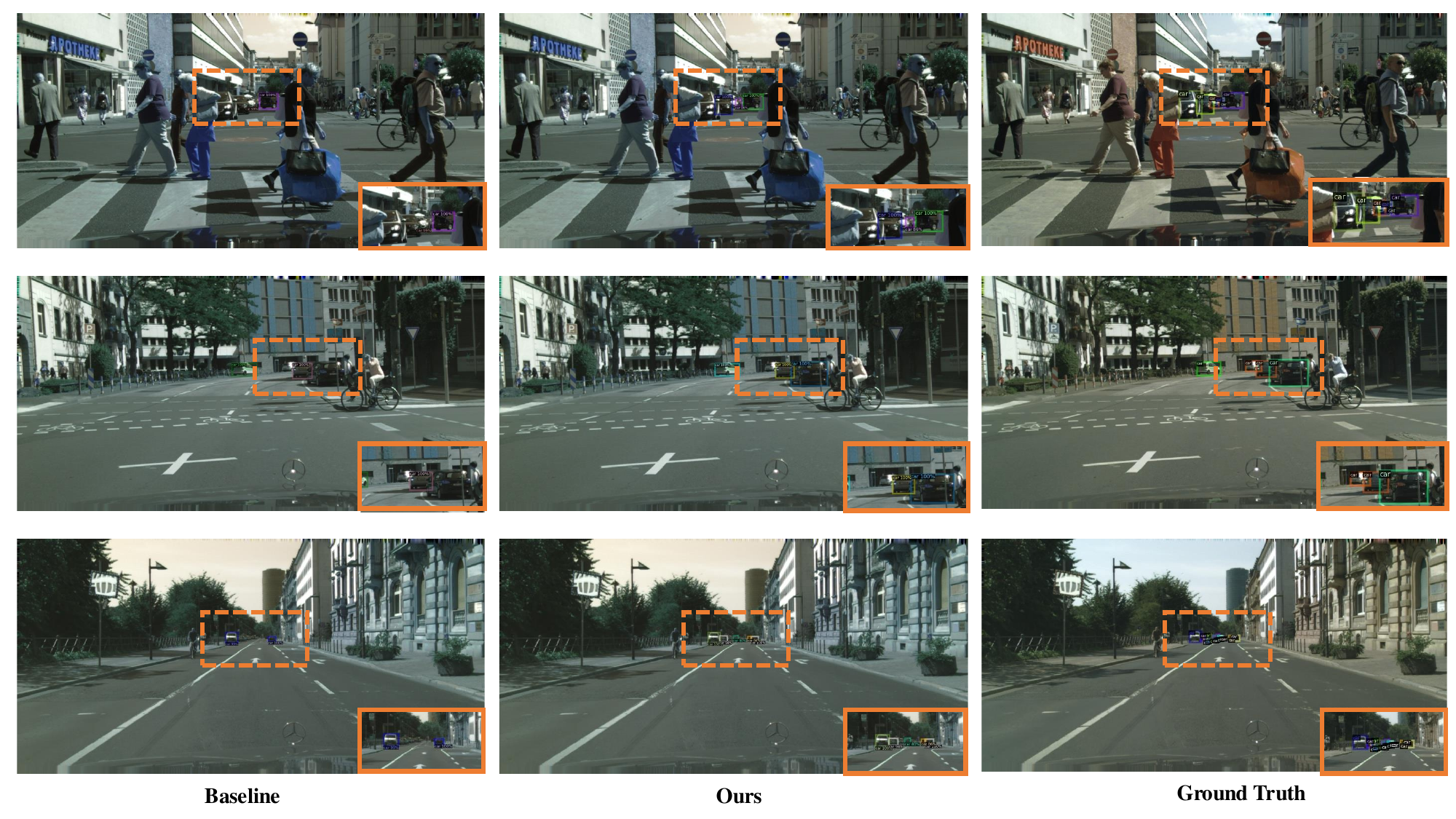}
    \caption{Qualitative results on adaptation from Sim10k to Cityscapes. Please zoom in for details.}
    \vspace{-2mm}
    \label{fig:vis-s2c}
\end{figure*}

We visualize the adaptation performance of the baseline model, our proposed model, and the ground truth on the Cityscapes to Foggy Cityscapes and Sim10k to Cityscapes tasks, respectively. As shown in Figures~\ref{fig:vis-c2f} and \ref{fig:vis-s2c}, our method significantly reduces the occurrence of false negative errors compared to the baseline model, demonstrating the effectiveness of our proposed approach in these domain adaptation scenarios. Additionally, it can be observed that our method improves the detection of objects surrounded by rich domain-specific information, such as those in regions with denser fog, further validating its robustness in challenging conditions.

\subsection{Limitation Analysis}
Although we have validated the effectiveness of our differential alignment strategy on the Faster-RCNN detector~\cite{1fasterrcnn}, further exploration on other detectors is still needed. For instance, it remains to be investigated whether the approach is effective when applied to deformable DETR~\cite{70deformable}. Additionally, we aim to explore more diverse adaptation scenarios in future work, such as domain adaptation from daytime to nighttime.

\end{document}